\documentclass[10pt,twocolumn,letterpaper]{article}

\usepackage{iccv}
\usepackage{times}

\iccvfinalcopy % *** Uncomment this line for the final submission

\usepackage{graphicx}
\usepackage{amsmath}
\usepackage{amssymb}
\usepackage{booktabs}
\usepackage{subfig}
\usepackage[font=small]{caption}

\usepackage{diagbox}
\usepackage{nicefrac}       %
\usepackage{microtype}      %

\usepackage{blindtext}
\usepackage{float}
\usepackage{enumitem}
\usepackage[table]{xcolor}
\usepackage{tabulary,multirow,overpic}
\usepackage{verbatim}
\usepackage{color}
\usepackage{multicol}
\usepackage{dblfloatfix}
\usepackage{pifont}%
\usepackage[accsupp]{axessibility}  %
\usepackage{makecell}

\usepackage{multicol}

\usepackage{fontawesome5}

\newlength\savewidth\newcommand\shline{\noalign{\global\savewidth\arrayrulewidth\global\arrayrulewidth 1pt}\hline\noalign{\global\arrayrulewidth\savewidth}}
\newcommand{\tablestyle}[2]{\setlength{\tabcolsep}{#1}\renewcommand{\arraystretch}{#2}\centering\footnotesize}

\newcolumntype{x}[1]{>{\centering\arraybackslash}p{#1pt}}
\newcolumntype{y}[1]{>{\raggedright\arraybackslash}p{#1pt}}
\newcolumntype{z}[1]{>{\raggedleft\arraybackslash}p{#1pt}}

\definecolor{baselinecolor}{gray}{.92}

\definecolor{demphcolor}{gray}{.2}

\definecolor{demphcolorinline}{gray}{.3}

\definecolor{demphcolor1}{gray}{.6}

\definecolor{eva01purple}{RGB}{168,119,200}
\newcommand{\evapurple}[1]{\textcolor{eva01purple}{#1}}

\definecolor{eva01green}{RGB}{82,208,83}
\newcommand{\evagreen}[1]{\textcolor{eva01green}{#1}}

\definecolor{eva02red}{RGB}{236,35,35}
\newcommand{\evared}[1]{\textcolor{eva02red}{#1}}

\definecolor{eva02yellow}{RGB}{249,157,83}

\definecolor{02pink}{RGB}{240,178,188}

\definecolor{00blue}{RGB}{100,149,237}
\newcommand{\evablue}[1]{\textcolor{00blue!80}{#1}}

\newcommand{\ph}[1]{\textcolor{white}{#1}}

\newcommand{\phgray}[1]{\textcolor{Graylight!30}{#1}}

\definecolor{citecolor}{RGB}{34,139,34}
\definecolor{citecolor2}{HTML}{0071bc}
\definecolor{Graylight}{gray}{0.9}
\definecolor{lightred}{RGB}{241,140,142}
\usepackage[pagebackref=true,breaklinks=true,colorlinks,
citecolor=eva02red,bookmarks=false]{hyperref}

\definecolor{clipbaselinecolor}{gray}{.9}

\definecolor{defaultcolor}{HTML}{E8E2F7}

\renewcommand{\paragraph}[1]{\vspace{1.25mm}\noindent\textbf{#1}}

\newcommand{\app}{\raise.17ex\hbox{$\scriptstyle\sim$}}
\newcommand{\appp}{\raise.20ex\hbox{$\scriptscriptstyle\sim$}}

\def\x{$\times$}

\newcommand{\tblref}[1]{Table~\ref{#1}}

\newcommand{\evaone}{{\textbf{\evapurple{EVA}}}\xspace}
\newcommand{\eva}{{\textbf{\evared{EVA-02}}}\xspace}
\newcommand{\evaOne}{{\textbf{\evapurple{EVA-01}}}\xspace}

\newcommand{\evaclip}{{\textbf{\evablue{EVA-CLIP}}}\xspace}
\newcommand{\evaOneclip}{{\textbf{\evablue{EVA-01-CLIP}}}\xspace}
\newcommand{\evaTwoclip}{{\textbf{\evablue{EVA-02-CLIP}}}\xspace}

\newcommand{\rgray}{\rowcolor{Graylight!30}}

\newcommand{\suptext}[1]{$^{\text{#1}}$}

\newcommand{\cmark}{\ding{51}}%
\newcommand{\xmark}{\ding{55}}%

\usepackage[capitalize]{cleveref}
\crefname{section}{Sec.}{Secs.}
\Crefname{section}{Section}{Sections}
\Crefname{table}{Table}{Tables}
\crefname{table}{Tab.}{Tabs.}
\newcommand{\authorskip}{\hspace{4mm}}

\begin{document}

\title{\evaclip: Improved Training Techniques for CLIP at Scale}

\author{
{
% \fontsize{11.pt}{9.84pt}\selectfont
Quan Sun\textsuperscript{1} \authorskip
Yuxin Fang\textsuperscript{1,2} \authorskip 
Ledell Wu\textsuperscript{1} \authorskip
Xinlong Wang\textsuperscript{1} \authorskip 
Yue Cao\textsuperscript{1}
}
\\[0.5mm]
{
\fontsize{10.0pt}{9.84pt}\selectfont
\textsuperscript{1}Beijing Academy of Artificial Intelligence \hspace{5.5mm} \textsuperscript{2}Huazhong University of Science and Technology 
}
\\[1mm]
{
\fontsize{8.4pt}{9.84pt}\selectfont
\textbf{Fight together with \href{https://en.wikipedia.org/wiki/Rei_Ayanami}{\color{00blue!80}Rei} at }\href{https://github.com/baaivision/EVA/tree/master/EVA-CLIP}{\color{00blue!80} \bfseries \ttfamily baaivision/EVA/CLIP}
}
}

\maketitle

\begin{abstract}
   Contrastive language-image pre-training, CLIP for short, has gained increasing attention for its potential in various scenarios. In this paper, we propose \evaclip, a series of models that significantly improve the efficiency and effectiveness of CLIP training. Our approach incorporates new techniques for representation learning, optimization, and augmentation, enabling \evaclip to achieve superior performance compared to previous CLIP models with the same number of parameters but significantly smaller training costs. Notably, our largest 5.0B-parameter \evaTwoclip-E/14+ with only 9 billion seen samples achieves \textbf{82.0}\% zero-shot top-1 accuracy on ImageNet-1K val. A smaller \evaTwoclip-L/14+ with only 430 million parameters and 6 billion seen samples achieves \textbf{80.4}\% zero-shot top-1 accuracy on ImageNet-1K val. To facilitate open access and open research, we release the complete suite of \evaclip to the community.
\end{abstract}

%   Contrastive language-image pre-training, CLIP for short, has gained increasing attention for its potential in various scenarios. In this paper, we propose EVA-CLIP, a series of models that significantly improve the efficiency and effectiveness of CLIP training. Our approach incorporates new techniques for representation learning, optimization, and augmentation, enabling EVA-CLIP to achieve superior performance compared to previous CLIP models with the same number of parameters but significantly smaller training costs. Notably, our largest 5.0B-parameter EVA-02-CLIP-E/14+ with only 9 billion seen samples achieves 82.0 zero-shot top-1 accuracy on ImageNet-1K val. A smaller EVA-02-CLIP-L/14+ with only 430 million parameters and 6 billion seen samples achieves 80.4 zero-shot top-1 accuracy on ImageNet-1K val. To facilitate open access and open research, we release the complete suite of EVA-CLIP to the community.

\section{Introduction}
CLIP (Contrastive Language-Image Pre-training) is a powerful vision-language foundation model that leverages large-scale datasets to learn rich visual representations by bridging vision and language via contrastive image-text pre-training. The CLIP model exhibits robust zero-shot transferability~\cite{clip}, and has the potential to enhance both multimodal and unimodal visual tasks, such as AI-generated content applications~\cite{dalle2,eva,blip2,laion5b}. Despite its significance, training CLIP models remains an inevitable challenge due to its high computational cost and training instability issues when scaling up. 

In this paper, we propose \evaclip, a family of models that provides a feasible, efficient, and effective solution for training CLIP models. Our approach incorporates several techniques that can significantly reduce training costs, stabilize the training process and improve zero-shot performance, including initialize CLIP with pre-trained EVA~\cite{eva, EVA02} representations, the LAMB~\cite{lamb} optimizer, randomly dropping input tokens~\cite{flip}, and a speedup trick named flash attention~\cite{flashattention}. With these techniques, we are able to greatly stabilize the training of CLIP models at scale with less computational costs and outperform the training-from-scratch counterpart with much fewer samples on a broad range of zero-shot benchmarks. Our largest 5.0B-parameter \evaTwoclip-E/14+ with only 9 billion seen samples achieves 82.0\% zero-shot top-1 accuracy on ImageNet-1K val. A smaller \evaTwoclip-L/14+ with only 430 million parameters and 6 billion seen samples achieves 80.4\% zero-shot top-1 accuracy on ImageNet-1K val.

\begin{figure}[t]
    \centering
    \vspace{3.5em}
    \includegraphics[width=1.0\linewidth]{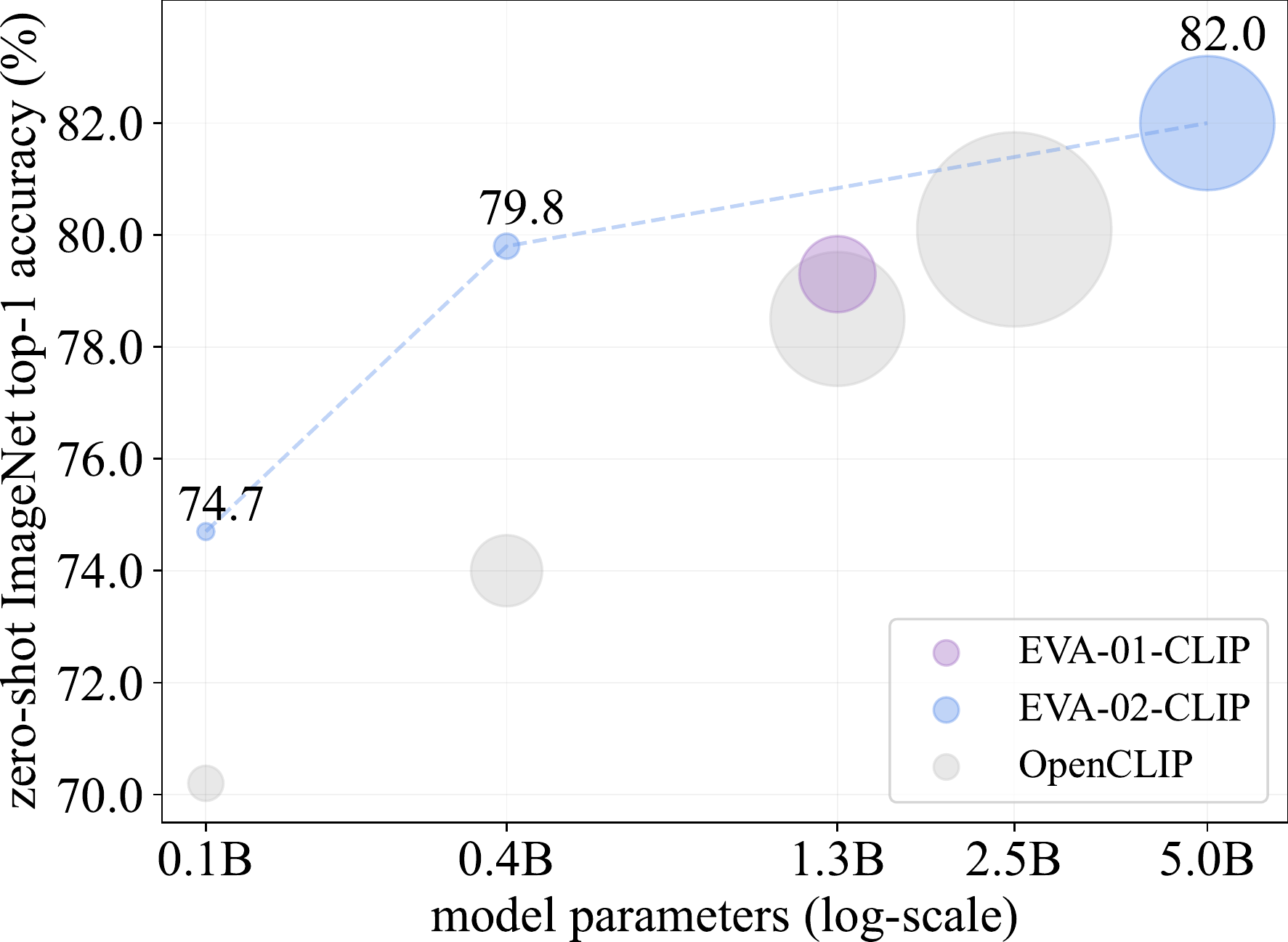}
    \caption{\textbf{Summary of CLIP models' ImageNet-1K zero-shot classification performance.} The diameter of each circle corresponds to forward GFLOPs x the number of training samples.}
    \label{fig:teaser}
\end{figure}

%#################################################
% clip
%#################################################
\begin{table*}[t!]
\vspace{-20pt}
\centering
%#################################################
% cfg
%#################################################
\subfloat[
\textbf{CLIP model configurations and zero-shot top-1 accuracy on ImageNet.} With all techniques, \evaOneclip-g/14 can be stably trained via pure $\mathtt{fp16}$ precision with fewer image-text pairs (11B \textit{v.s.} 32B) sampled on \app1/7\x GPUs compared with Open CLIP-H/14. \evaOneclip-g/14 with smaller text encoder (124M \textit{v.s.} 354M) was trained on a smaller dataset (LAION-400M \textit{v.s.} LAION-2B) and only used the $\mathtt{DeepSpeed}$ optimization library~\cite{rasley2020deepspeed} with ZeRO stage-1 optimizer~\cite{rajbhandari2020zero}. \evaTwoclip-E/14 with smaller text encoder (354M \textit{v.s.} 695M) was trained with pure $\mathtt{fp16}$.
\label{tab: clip cfg}
]{
\centering
\begin{minipage}{1\linewidth}{\begin{center}
    \tablestyle{3.5pt}{1.2}
    \begin{tabular}{r|c|ccc|cc|ccc|c}
        & & total & image & text & & & & & & \scriptsize{IN-1K} \\
        model{\scriptsize{\ph{+}}} & precision & \#param. & \#param. & \#param. & data & samples seen & image size & batch size & gpus for training & zs top-1 \\
        \shline
        \multicolumn{11}{c}{\scriptsize (a) comparisons with CLIP-\textbf{Base} baselines} \\
        \hline
        \scriptsize OpenAI CLIP-B/16\ph{+} & $\mathtt{float16}$ & \scriptsize 149M & \scriptsize 86M & \scriptsize 63M & \scriptsize CLIP-400M & \scriptsize 13B & \scriptsize 224\suptext{2} & \scriptsize 32k & \scriptsize {unknown} & \scriptsize 68.3 \\
        \rgray
        \scriptsize {\evaOneclip-B/16\phgray{+}} & $\mathtt{float16}$ & \scriptsize 149M & \scriptsize 86M & \scriptsize 63M & \scriptsize Merged-2B & \scriptsize 3B & \scriptsize 224\suptext{2} & \scriptsize 32k & \scriptsize 16{\scriptsize{\x}}A100 \scriptsize{(40GB)} & \scriptsize 69.7 \\
        \scriptsize Open CLIP-B/16\ph{+} & $\mathtt{float16}$ & \scriptsize 149M & \scriptsize 86M & \scriptsize 63M & \scriptsize LAION-2B & \scriptsize 34B & \scriptsize 224\suptext{2} & \scriptsize 88k & \scriptsize 344{\scriptsize{\x}}A100 \scriptsize{(40GB)} & \scriptsize 70.2 \\
        \rgray
        \scriptsize {\evaTwoclip-B/16\phgray{+}} & $\mathtt{float16}$ & \scriptsize 149M & \scriptsize 86M & \scriptsize 63M & \scriptsize Merged-2B & \scriptsize \textbf{8B} & \scriptsize 224\suptext{2} & \scriptsize 131k & \scriptsize \textbf{64{\scriptsize{\x}}A100 \scriptsize{(40GB)}} &  \scriptsize \textbf{74.7} \\
        \shline
        \multicolumn{11}{c}{\scriptsize (b) comparisons with CLIP-\textbf{Large} baselines} \\
        \hline
        \scriptsize Open CLIP-L/14\ph{+} & $\mathtt{float16}$ & \scriptsize 428M & \scriptsize 304M & \scriptsize 124M & \scriptsize LAION-2B & \scriptsize 32B & \scriptsize 224\suptext{2} & \scriptsize 79k & \scriptsize 384{\scriptsize{\x}}A100 \scriptsize{(40GB)} & \scriptsize 74.0 \\
        \scriptsize OpenAI CLIP-L/14\ph{+} & $\mathtt{float16}$ & \scriptsize 428M & \scriptsize 304M & \scriptsize 124M & \scriptsize CLIP-400M & \scriptsize 13B & \scriptsize 224\suptext{2} & \scriptsize 32k & \scriptsize 256{\scriptsize{\x}V100 \scriptsize{(32GB)}} & \scriptsize 75.5 \\
        \rgray
        \scriptsize {\evaTwoclip-L/14\phgray{+}} & $\mathtt{float16}$ & \scriptsize 428M & \scriptsize 304M & \scriptsize 124M & \scriptsize Merged-2B & \scriptsize \textbf{4B} & \scriptsize 224\suptext{2} & \scriptsize 131k & \scriptsize \textbf{128{\scriptsize{\x}}A100} \scriptsize \scriptsize{(40GB)} & \scriptsize \textbf{79.8} \\
        \shline
        \multicolumn{11}{c}{\scriptsize (c) comparisons with \textbf{larger} CLIPs trained with \textbf{more samples}} \\
        \hline
        \scriptsize OpenAI CLIP-L/14+ & $\mathtt{float16}$ & \scriptsize 428M & \scriptsize 304M & \scriptsize 124M & \scriptsize CLIP-400M & \scriptsize 13B & \scriptsize 336\suptext{2} & \scriptsize 32k & \scriptsize 256{\scriptsize{\x}V100 \scriptsize{(32GB)}} & \scriptsize 76.6 \\
        
        \scriptsize Open CLIP-H/14\ph{+} & $\mathtt{bfloat16}$ & \scriptsize 1.0B & \scriptsize 632M & \scriptsize 354M & \scriptsize LAION-2B & \scriptsize 32B & \scriptsize 224\suptext{2} & \scriptsize 79k & \scriptsize 824{\scriptsize{\x}}A100 \scriptsize{(40GB)} & \scriptsize 78.0 \\
        \scriptsize Open CLIP-g/14\phgray{+} & $\mathtt{bfloat16}$ & \scriptsize 1.3B & \scriptsize 1.0B & \scriptsize 354M & \scriptsize LAION-2B & \scriptsize 34B & \scriptsize 224\suptext{2} & \scriptsize 88k & \scriptsize unknown & \scriptsize 78.5 \\
        
        % \hline
        \rgray
        \scriptsize {\evaOneclip-g/14\phgray{+}} & $\mathtt{float16}$ & \scriptsize 1.1B & \scriptsize 1.0B & \scriptsize 124M & \scriptsize LAION-400M & \scriptsize 11B & \scriptsize 224\suptext{2} & \scriptsize 41k & \scriptsize 256{\scriptsize{\x}}A100 \scriptsize{(40GB)} & \scriptsize 78.5 \\
        \rgray
        \scriptsize {\evaOneclip-g/14+} & $\mathtt{float16}$ & \scriptsize 1.3B & \scriptsize 1.0B & \scriptsize 354M & \scriptsize Merged-2B & \scriptsize 11B & \scriptsize 224\suptext{2} & \scriptsize 114k & \scriptsize 112{\scriptsize{\x}}A100 \scriptsize{(40GB)} & \scriptsize 79.3 \\
        \scriptsize Open CLIP-G/14\ph{+} & $\mathtt{bfloat16}$ & \scriptsize 2.5B & \scriptsize 1.8B & \scriptsize 695M & \scriptsize LAION-2B & \scriptsize 39B & \scriptsize 224\suptext{2} & \scriptsize 160k & \scriptsize 512{\scriptsize{\x}}A100 \scriptsize{(80GB)} & \scriptsize 80.1 \\
        \rgray
        \scriptsize {\evaTwoclip-L/14+} & $\mathtt{float16}$ & \scriptsize 428M & \scriptsize 304M & \scriptsize 124M & \scriptsize Merged-2B & \scriptsize 6B & \scriptsize 336\suptext{2} & \scriptsize 61k & \scriptsize 128{\scriptsize{\x}}A100 \scriptsize{(40GB)} & \scriptsize 80.4 \\
        \rgray
        \scriptsize {\evaTwoclip-E/14\phgray{+}} & $\mathtt{float16}$ & \scriptsize 4.7B & \scriptsize 4.4B & \scriptsize 354M & \scriptsize LAION-2B & \scriptsize 4B & \scriptsize 224\suptext{2} & \scriptsize 115k & \scriptsize 144{\scriptsize{\x}}A100 \scriptsize{(80GB)} & \scriptsize 81.9 \\
        \rgray
        \scriptsize {\evaTwoclip-E/14+} & $\mathtt{bfloat16}$ & \scriptsize 5.0B & \scriptsize 4.4B & \scriptsize 695M & \scriptsize LAION-2B & \scriptsize \textbf{9B} & \scriptsize 224\suptext{2} & \scriptsize 144k & \scriptsize \textbf{144{\scriptsize{\x}}A100 \scriptsize{(80GB)}} & \scriptsize \textbf{82.0} \\
        \end{tabular}
\end{center}}\end{minipage}
}
\vspace{0.5em}
\\

%#################################################
% clip result
%#################################################
\subfloat[
\textbf{Summary of zero-shot performance on ImageNet variants and ObjectNet.} ``avg. acc.'': the averaged top-1 accuracy on different ImageNet variants (\ie, IN-\{1K, V2, ReaL, Adv., Ren., Ske.\}), and ObjectNet. ``{$\Delta$\scriptsize{$\downarrow$}}'': The gap between the averaged top-1 accuracy and the ImageNet-1K top-1 accuracy (the lower the better). \evaclip suffers from the smallest performance drop (only \textbf{\evagreen{0.8}}\% top-1 accuracy gap for \evaTwoclip-L/14+ and \textbf{\evagreen{1.1}}\% for \evaTwoclip-E/14+) while \evaTwoclip-E/14+ achieves the highest zero-shot classification accuracy averaged on all 6 benchmarks (\textbf{\evagreen{80.9}}\% averaged top-1 accuracy).
\label{tab: clip result}
]{
\centering
\begin{minipage}{1\linewidth}{\begin{center}
\tablestyle{3.5pt}{1.2}
    \begin{tabular}{r|cccccc|c|c}
        method\ph{+} & \scriptsize IN-1K & \scriptsize IN-A & \scriptsize IN-R & \scriptsize IN-V2 & \scriptsize IN-Sketch & \scriptsize ObjectNet & $\Delta$\scriptsize{$\downarrow$} & \textbf{avg. acc.} \\
        \shline
        \multicolumn{9}{c}{\scriptsize (a) comparisons with CLIP-\textbf{Base} baselines} \\
        \hline
        \scriptsize OpenAI CLIP-B/16\ph{+} & \scriptsize 68.3 & \scriptsize 50.0 & \scriptsize 77.7 & \scriptsize 61.9 & \scriptsize 48.2 & \scriptsize 55.3 & \scriptsize \textbf{8.1} & \scriptsize 60.2 \\
        \scriptsize Open CLIP-B/16\ph{+} & \scriptsize 70.2 & \scriptsize 38.2 & \scriptsize 80.6 & \scriptsize 62.3 & \scriptsize 56.1 & \scriptsize 56.0 & \scriptsize{9.6} & \scriptsize 60.6 \\
        \rgray
        \scriptsize \evaTwoclip-B/16\phgray{+} & \scriptsize \textbf{74.7} & \scriptsize \textbf{54.1} & \scriptsize \textbf{82.5} & \scriptsize \textbf{67.0} & \scriptsize \textbf{57.7} & \scriptsize \textbf{62.3}  & \scriptsize {8.3} & \scriptsize \textbf{66.4} \\
        \shline
        \multicolumn{9}{c}{\scriptsize (b) comparisons with CLIP-\textbf{Large} baselines} \\
        \hline
        \scriptsize Open CLIP-L/14\ph{+} & \scriptsize 74.0 & \scriptsize 48.0 & \scriptsize 86.5 & \scriptsize 66.4 & \scriptsize 61.8 & \scriptsize 61.1 & \scriptsize{7.7} & \scriptsize 66.3 \\
        \scriptsize OpenAI CLIP-L/14\ph{+} & \scriptsize 75.5 & \scriptsize 70.8 & \scriptsize 87.8 & \scriptsize 69.9 & \scriptsize 59.6 & \scriptsize 69.0 & \scriptsize{3.4} & \scriptsize 72.1 \\
        \rgray
        \scriptsize \evaTwoclip-L/14\phgray{+} & \scriptsize \textbf{79.8} & \scriptsize \textbf{76.1} & \scriptsize \textbf{92.7} & \scriptsize \textbf{72.9} & \scriptsize \textbf{68.1} & \scriptsize \textbf{75.3}  & \scriptsize \textbf{2.9} & \scriptsize \textbf{77.5} \\
        \shline
        \multicolumn{9}{c}{\scriptsize (c) comparisons with \textbf{larger} CLIPs trained with \textbf{more samples}} \\
        \hline
        \scriptsize Open CLIP-H/14\ph{+} & \scriptsize 78.0 & \scriptsize 59.3 & \scriptsize 89.3 & \scriptsize 70.9 & \scriptsize 66.6 & \scriptsize 69.7 & \scriptsize{5.7} & \scriptsize 72.3\\
        \scriptsize Open CLIP-g/14\ph{+} & \scriptsize 78.5 & \scriptsize 60.8 & \scriptsize 90.2 & \scriptsize 71.7 & \scriptsize 67.5 & \scriptsize 69.2 & \scriptsize{5.5} & \scriptsize 73.0 \\
        \scriptsize OpenAI CLIP-L/14+ & \scriptsize 76.6 & \scriptsize 77.5 & \scriptsize 89.0 & \scriptsize 70.9 & \scriptsize 61.0 & \scriptsize 72.0 & \scriptsize{2.1} & \scriptsize 74.5 \\
        \rgray
        \scriptsize \evaOneclip-g/14\phgray{+} & \scriptsize 78.5 & \scriptsize 73.6 & \scriptsize 92.5 & \scriptsize 71.5 & \scriptsize 67.3 & \scriptsize 72.3 & \scriptsize{2.5} & \scriptsize 76.0 \\
        \scriptsize Open CLIP-G/14\ph{+} & \scriptsize 80.1 & \scriptsize 69.3 & \scriptsize 92.1 & \scriptsize 73.6 & \scriptsize 68.9 & \scriptsize 73.0 & \scriptsize{3.9} & \scriptsize 76.2\\
        \rgray
        \scriptsize \evaOneclip-g/14+ & \scriptsize 79.3 & \scriptsize 74.1 & \scriptsize 92.5 & \scriptsize 72.1 & \scriptsize 68.1 & \scriptsize 75.3 & \scriptsize{2.4} & \scriptsize 76.9\\
        \rgray
        \scriptsize \evaTwoclip-L/14+ & \scriptsize 80.4 & \scriptsize \textbf{82.9} & \scriptsize 93.2 & \scriptsize 73.8 & \scriptsize 68.9 & \scriptsize 78.4 & \scriptsize \textbf{0.8} & \scriptsize 79.6 \\
        \rgray
        \scriptsize \evaTwoclip-E/14\phgray{+} & \scriptsize 81.9 & \scriptsize 80.4 & \scriptsize 94.1 & \scriptsize 75.4 & \scriptsize 71.8 & \scriptsize 76.9 & \scriptsize 1.8 & \scriptsize 80.1\\
        \rgray
        \scriptsize \evaTwoclip-E/14+ & \scriptsize \textbf{82.0} & \scriptsize 82.1 & \scriptsize \textbf{94.5} & \scriptsize \textbf{75.7} & \scriptsize \textbf{71.6} & \scriptsize \textbf{79.6} & \scriptsize 1.1 & \scriptsize \textbf{80.9}\\
    \end{tabular}
\end{center}}\end{minipage}
}
\vspace{-.3em}
\caption{\textbf{CLIP model configurations and highlights of our key results.} Shown are the top-1 accuracy of our method, \evaclip, and similar baselines – CLIP and Open CLIP – on ImageNet and other robustness test sets. None of these models have learned from any labeled training example in ImageNet.}
\label{tab: clip config and results}
% \vspace{-10pt}
\end{table*}
%##################################################################################################

%#################################################
% image cls & retrieval
%#################################################
\begin{table*}[t!]
\vspace{-10pt}
\centering
\tablestyle{1.4pt}{1.2}
    \begin{tabular}{r|cccccc|ccccccccccccccccccccc|c}
        % \rotatebox[origin=l]{90}{datasets} 
        &
        \rotatebox[origin=l]{90}{\scriptsize{ImageNet-1K~\cite{deng2009imagenet}}} &
        \rotatebox[origin=l]{90}{\scriptsize{ImageNet-V2~\cite{recht2019imagenetv2}}} &
        \rotatebox[origin=l]{90}{\scriptsize{ImageNet-Adv.~\cite{inadv}}} &
        \rotatebox[origin=l]{90}{\scriptsize{ImageNet-Ren.~\cite{inren}}} &
        \rotatebox[origin=l]{90}{\scriptsize{ImageNet-Ske.~\cite{inske}}} &
        \rotatebox[origin=l]{90}{\scriptsize{ObjectNet~\cite{objectnet}}} &
        \rotatebox[origin=l]{90}{\scriptsize{CIFAR-10~\cite{cifar}}} &
        \rotatebox[origin=l]{90}{\scriptsize{CIFAR-100~\cite{cifar}}} & 
        \rotatebox[origin=l]{90}{\scriptsize{MNIST~\cite{lecun1998gradient}}} & 
        \rotatebox[origin=l]{90}{\scriptsize{Caltech-101~\cite{fei2004learning}}} & 
        \rotatebox[origin=l]{90}{\scriptsize{SUN397~\cite{xiao2010sun}}} & 
        \rotatebox[origin=l]{90}{\scriptsize{FGVC Aircraft~\cite{maji2013fine}}} & 
        \rotatebox[origin=l]{90}{\scriptsize{Country-211~\cite{clip}}} & 
        \rotatebox[origin=l]{90}{\scriptsize{Stanford Cars~\cite{krause20133d}}} &
        \rotatebox[origin=l]{90}{\scriptsize{Birdsnap~\cite{berg2014birdsnap}}} & 
        \rotatebox[origin=l]{90}{\scriptsize{DTD~\cite{cimpoi14describing}}} & 
        \rotatebox[origin=l]{90}{\scriptsize{Eurosat~\cite{helber2019eurosat}}} & 
        \rotatebox[origin=l]{90}{\scriptsize{FER2013~\cite{goodfellow2013challenges}}} & 
        \rotatebox[origin=l]{90}{\scriptsize{Flowers-102~\cite{nilsback2008automated}}} & 
        \rotatebox[origin=l]{90}{\scriptsize{Food-101~\cite{bossard2014food}}} & 
        \rotatebox[origin=l]{90}{\scriptsize{GTSRB~\cite{stallkamp2012man}}} & 
        \rotatebox[origin=l]{90}{\scriptsize{PCam~\cite{veeling2018rotation}}} & 
        \rotatebox[origin=l]{90}{\scriptsize{Pets~\cite{parkhi12a}}} & 
        \rotatebox[origin=l]{90}{\scriptsize{Rendered SST2~\cite{clip}}} & 
        \rotatebox[origin=l]{90}{\scriptsize{Resisc45~\cite{cheng2017remote}}} & 
        \rotatebox[origin=l]{90}{\scriptsize{STL10~\cite{coates2011analysis}}} & 
        \rotatebox[origin=l]{90}{\scriptsize{VOC2017~\cite{everingham2015pascal}}} &
        \rotatebox[origin=l]{90}{\ph{.}\textbf{avg. top-1 acc.}}
        \\
        \shline

        \multicolumn{29}{c}{\scriptsize (a) comparisons with CLIP-\textbf{Base} baselines} \\
        \hline
        \scriptsize OpenAI CLIP-B/16\ph{+} & \scriptsize 68.3 & \scriptsize 61.9 & \scriptsize 50.0 & \scriptsize 77.7 & \scriptsize 48.2 & \scriptsize 55.3 & \scriptsize 90.8 & \scriptsize 67.0 & \scriptsize 51.6 & \scriptsize 84.7 & \scriptsize 64.4 & \scriptsize 24.4 & \scriptsize \textbf{22.8} & \scriptsize 64.8 & \scriptsize 34.5 & \scriptsize 44.7 & \scriptsize 55.0 & \scriptsize 46.2 & \scriptsize 71.3 & \scriptsize 88.8 & \scriptsize 43.5 & \scriptsize 50.7 & \scriptsize 89.1 & \scriptsize \textbf{60.8} & \scriptsize 59.1 & \scriptsize 98.3 & \scriptsize 78.3 & \scriptsize 61.2  \\
        \scriptsize Open CLIP-B/16\ph{+} & \scriptsize 70.2 & \scriptsize 62.3 & \scriptsize 38.2 & \scriptsize 80.6 & \scriptsize 56.1 & \scriptsize 56.0 & \scriptsize 94.9 & \scriptsize 76.9 & \scriptsize \textbf{65.9} & \scriptsize \textbf{86.7} & \scriptsize \textbf{70.8} & \scriptsize \textbf{27.0} & \scriptsize 20.3 & \scriptsize \textbf{88.5} & \scriptsize \textbf{42.1} & \scriptsize \textbf{56.6} & \scriptsize 52.7 & \scriptsize \textbf{51.8} & \scriptsize 71.4 & \scriptsize 86.6 & \scriptsize \textbf{48.3} & \scriptsize \textbf{56.3} & \scriptsize 90.3 & \scriptsize 60.0 & \scriptsize 63.4 & \scriptsize 97.9 & \scriptsize 70.8 & \scriptsize 64.5 \\
        \rgray
        \scriptsize \evaTwoclip-B/16\phgray{+} & \scriptsize \textbf{74.7} & \scriptsize \textbf{67.0} & \scriptsize \textbf{54.1} & \scriptsize \textbf{82.5} & \scriptsize \textbf{57.7} & \scriptsize \textbf{62.3} & \scriptsize \textbf{98.4} & \scriptsize \textbf{87.7} & \scriptsize 47.9 & \scriptsize 86.3 & \scriptsize 70.7 & \scriptsize 24.8 & \scriptsize 21.4 & \scriptsize 78.6 & \scriptsize 37.7 & \scriptsize 53.1 & \scriptsize \textbf{67.0} & \scriptsize 51.2 & \scriptsize \textbf{75.9} & \scriptsize \textbf{89.4} & \scriptsize 46.3 & \scriptsize 50.9 & \scriptsize \textbf{92.2} & \scriptsize 54.1 & \scriptsize \textbf{60.7} & \scriptsize \textbf{99.5} & \scriptsize \textbf{80.2} & \scriptsize \textbf{65.6}  \\
        \shline
        
        \multicolumn{29}{c}{\scriptsize (b) comparisons with CLIP-\textbf{Large} baselines} \\
        \hline
        \scriptsize OpenAI CLIP-L/14\ph{+} & \scriptsize 75.5 & \scriptsize 69.9 & \scriptsize 70.8 & \scriptsize 87.8 & \scriptsize 59.6 & \scriptsize 69.0 & \scriptsize 95.6 & \scriptsize 75.8 & \scriptsize 76.4 & \scriptsize 86.7 & \scriptsize 67.6 & \scriptsize 31.4 & \scriptsize \textbf{31.9} & \scriptsize 77.9 & \scriptsize 40.5 & \scriptsize 55.4 & \scriptsize 62.4 & \scriptsize 49.9 & \scriptsize \textbf{79.2} & \scriptsize 93.1 & \scriptsize 50.6 & \scriptsize 52.0 & \scriptsize 93.5 & \scriptsize \textbf{68.9} & \scriptsize 64.6 & \scriptsize 99.4 & \scriptsize 67.6 & \scriptsize 69.0 \\
        \scriptsize Open CLIP-L/14\ph{+} & \scriptsize 74.0 & \scriptsize 66.4 & \scriptsize 48.0 & \scriptsize 86.5 & \scriptsize 61.8 & \scriptsize 61.1 & \scriptsize 95.8 & \scriptsize 78.4 & \scriptsize 64.4 & \scriptsize 87.6 & \scriptsize 74.0 & \scriptsize 34.8 & \scriptsize 24.4 & \scriptsize \textbf{91.5} & \scriptsize \textbf{45.2} & \scriptsize 61.5 & \scriptsize 57.8 & \scriptsize 50.9 & \scriptsize 74.4 & \scriptsize 88.8 & \scriptsize 51.6 & \scriptsize \textbf{55.5} & \scriptsize 92.8 & \scriptsize 60.3 & \scriptsize 67.3 & \scriptsize 98.5 & \scriptsize 74.0 & \scriptsize 67.9 \\
        \rgray
        \scriptsize \evaTwoclip-L/14\phgray{+} & \scriptsize \textbf{79.8} & \scriptsize \textbf{72.9} & \scriptsize \textbf{76.1} & \scriptsize \textbf{92.7} & \scriptsize \textbf{68.1} & \scriptsize \textbf{75.3} & \scriptsize \textbf{99.1} & \scriptsize \textbf{90.7} & \scriptsize \textbf{67.3} & \scriptsize \textbf{88.3} & \scriptsize \textbf{74.1} & \scriptsize \textbf{36.2} & \scriptsize 30.9 & \scriptsize 90.5 & \scriptsize 43.8 & \scriptsize \textbf{63.2} & \scriptsize \textbf{70.0} & \scriptsize \textbf{53.4} & \scriptsize 77.0 & \scriptsize \textbf{93.4} & \scriptsize \textbf{57.1} & \scriptsize 54.2 & \scriptsize \textbf{93.9} & \scriptsize 61.6 & \scriptsize \textbf{70.0} & \scriptsize \textbf{99.6} & \scriptsize \textbf{82.2} & \scriptsize \textbf{72.6} \\
        \shline
        
        \multicolumn{29}{c}{\scriptsize (c) comparisons with \textbf{larger} CLIPs trained with \textbf{more samples}} \\
        \hline
        \scriptsize OpenAI CLIP-L/14+ & \scriptsize 76.6 & \scriptsize 70.9 & \scriptsize 77.5 & \scriptsize 89.0 & \scriptsize 61.0 & \scriptsize 72.0 & \scriptsize 94.9 & \scriptsize 74.4 & \scriptsize \textbf{79.0} & \scriptsize 87.2 & \scriptsize 68.7 & \scriptsize 33.4 & \scriptsize 34.5 & \scriptsize 79.3 & \scriptsize 41.0 & \scriptsize 56.0 & \scriptsize 61.5 & \scriptsize 49.1 & \scriptsize 78.6 & \scriptsize 93.9 & \scriptsize 52.4 & \scriptsize 60.8 & \scriptsize 93.8 & \scriptsize \textbf{70.7} & \scriptsize 65.4 & \scriptsize 99.4 & \scriptsize 78.1 & \scriptsize 70.3 \\
        \rgray
        \scriptsize \evaOneclip-g/14\phgray{+} & \scriptsize 78.5 & \scriptsize 71.5 & \scriptsize 73.6 & \scriptsize 92.5 & \scriptsize 67.3 & \scriptsize 72.3 & \scriptsize 98.3 & \scriptsize 88.7 & \scriptsize 62.3 & \scriptsize 87.7 & \scriptsize 74.2 & \scriptsize 32.4 & \scriptsize 28.6 & \scriptsize 91.7 & \scriptsize 50.0 & \scriptsize 61.3 & \scriptsize 73.6 & \scriptsize 52.2 & \scriptsize 74.5 & \scriptsize 93.5 & \scriptsize 49.1 & \scriptsize 49.9 & \scriptsize 94.2 & \scriptsize 58.4 & \scriptsize 70.3 & \scriptsize 98.9 & \scriptsize 83.2 & \scriptsize 71.4 \\
        \scriptsize Open CLIP-g/14\ph{+} & \scriptsize 78.5 & \scriptsize 71.7 & \scriptsize 60.8 & \scriptsize 90.2 & \scriptsize 67.5 & \scriptsize 69.2 & \scriptsize 98.2 & \scriptsize 84.7 & \scriptsize 71.9 & \scriptsize 88.1 & \scriptsize 74.1 & \scriptsize 44.6 & \scriptsize 30.9 & \scriptsize 94.0 & \scriptsize 51.0 & \scriptsize 68.7 & \scriptsize 64.7 & \scriptsize 55.8 & \scriptsize 81.0 & \scriptsize 92.4 & \scriptsize 49.7 & \scriptsize 53.8 & \scriptsize 93.9 & \scriptsize 56.7 & \scriptsize 69.6 & \scriptsize 98.9 & \scriptsize 81.6 & \scriptsize 71.9 \\
        
        \scriptsize Open CLIP-H/14\ph{+} & \scriptsize 78.0 & \scriptsize 70.8 & \scriptsize 59.2 & \scriptsize 89.3 & \scriptsize 66.6 & \scriptsize 69.7 & \scriptsize 97.4 & \scriptsize 84.7 & \scriptsize 72.9 & \scriptsize 85.0 & \scriptsize 75.2 & \scriptsize 42.8 & \scriptsize 30.0 & \scriptsize 93.5 & \scriptsize 52.9 & \scriptsize 67.8 & \scriptsize 72.7 & \scriptsize 52.0 & \scriptsize 80.1 & \scriptsize 92.7 & \scriptsize 58.4 & \scriptsize 54.2 & \scriptsize 94.5 & \scriptsize 64.3 & \scriptsize 70.5 & \scriptsize 98.5 & \scriptsize 77.7 & \scriptsize 72.3 \\
        \rgray
        \scriptsize \evaTwoclip-L/14+ & \scriptsize{80.4} & \scriptsize{73.8} & \scriptsize{\textbf{82.9}} & \scriptsize{93.2} & \scriptsize{68.9} & \scriptsize{78.4} & \scriptsize{98.9} & \scriptsize{89.8} & \scriptsize{64.3} & \scriptsize{89.5} & \scriptsize{74.8} & \scriptsize{37.5} & \scriptsize{33.6} & \scriptsize{91.6} & \scriptsize 45.8 & \scriptsize 64.5 & \scriptsize 71.4 & \scriptsize 51.0 & \scriptsize 77.2 & \scriptsize {94.2} & \scriptsize 57.6 & \scriptsize 54.9 & \scriptsize 94.2 & \scriptsize 64.6 & \scriptsize 69.8 & \scriptsize \textbf{99.7} & \scriptsize {82.7} & \scriptsize {73.5} \\
        \rgray
        \scriptsize \evaOneclip-g/14+ & \scriptsize 79.3 & \scriptsize 72.1 & \scriptsize 74.1 & \scriptsize 92.5 & \scriptsize 68.1 & \scriptsize 75.3 & \scriptsize 99.1 & \scriptsize 90.1 & \scriptsize 71.8 & \scriptsize 88.1 & \scriptsize 74.3 & \scriptsize 39.4 & \scriptsize 30.8 & \scriptsize 90.7 & \scriptsize 52.6 & \scriptsize 67.3 & \scriptsize 73.2 & \scriptsize 56.0 & \scriptsize 79.7 & \scriptsize 93.7 & \scriptsize 66.5 & \scriptsize 62.3 & \scriptsize 94.8 & \scriptsize 58.6 & \scriptsize 71.4 & \scriptsize {99.5} & \scriptsize 82.9 & \scriptsize 74.2 \\
        \scriptsize Open CLIP-G/14\ph{+} & \scriptsize 80.1 & \scriptsize 73.6 & \scriptsize 69.3 & \scriptsize 92.1 & \scriptsize 68.9 & \scriptsize 73.0 & \scriptsize 98.2 & \scriptsize 87.5 & \scriptsize 71.6 & \scriptsize 86.4 & \scriptsize 74.5 & \scriptsize 49.7 & \scriptsize 33.8 & \scriptsize 94.5 & \scriptsize 54.5 & \scriptsize \textbf{69.0} & \scriptsize 70.0 & \scriptsize \textbf{59.5} & \scriptsize 81.5 & \scriptsize 93.1 & \scriptsize 62.5 & \scriptsize 63.6 & \scriptsize 95.2 & \scriptsize 65.2 & \scriptsize 72.6 & \scriptsize 98.5 & \scriptsize 80.7 & \scriptsize 74.8 \\
        \rgray
        \scriptsize \evaTwoclip-E/14\phgray{+} & \scriptsize 81.9 & \scriptsize 75.4 & \scriptsize {80.4} & \scriptsize 94.1 & \scriptsize \textbf{71.8} & \scriptsize 76.9 & \scriptsize \textbf{99.3} & \scriptsize 92.5 & \scriptsize 76.7 & \scriptsize 89.0 & \scriptsize \textbf{76.5} & \scriptsize 47.9 & \scriptsize 34.7 & \scriptsize 94.4 & \scriptsize 56.3 & \scriptsize 68.2 & \scriptsize \textbf{77.6} & \scriptsize 55.1 & \scriptsize 82.5 & \scriptsize \textbf{95.2} & \scriptsize 67.1 & \scriptsize 49.6 & \scriptsize 95.6 & \scriptsize 61.1 & \scriptsize 73.5 & \scriptsize 99.2 & \scriptsize 83.0 & \scriptsize 76.1 \\
        \rgray
        \scriptsize \evaTwoclip-E/14+ & \scriptsize \textbf{82.0} & \scriptsize \textbf{75.7} & \scriptsize {82.1} & \scriptsize \textbf{94.5} & \scriptsize 71.6 & \scriptsize \textbf{79.6} & \scriptsize \textbf{99.3} & \scriptsize \textbf{93.1} & \scriptsize 74.7 & \scriptsize \textbf{90.5} & \scriptsize 75.1 & \scriptsize \textbf{54.1} & \scriptsize \textbf{35.7} & \scriptsize \textbf{94.6} & \scriptsize \textbf{58.1} & \scriptsize 68.2 & \scriptsize 75.8 & \scriptsize 58.6 & \scriptsize \textbf{84.5} & \scriptsize 94.9 & \scriptsize \textbf{67.7} & \scriptsize \textbf{63.7} & \scriptsize \textbf{95.8} & \scriptsize 61.4 & \scriptsize \textbf{75.6} & \scriptsize 99.2 & \scriptsize \textbf{85.6} & \scriptsize \textbf{77.5} \\
        \end{tabular}
\vspace{-1.em}
\caption{\textbf{Summary of \evaclip zero-shot image classification performance on 27 datasets.}}
\label{tab: clip zs img cls 27}
% \vspace{-2.5em}
\end{table*}

%#################################################
% video classification
%#################################################
\begin{table}[h]
% \vspace{-1.em}
\centering
    \tablestyle{1.5pt}{1.2}
    \begin{tabular}{r|x{27}x{20}x{20}x{20}|c}
        method\ph{+} & \scriptsize UCF-101 & \scriptsize K-400 & \scriptsize K-600 & \scriptsize K-700 & \textbf{avg. acc.} \\
        \shline
        \multicolumn{6}{c}{\scriptsize (a) comparisons with CLIP-\textbf{Base} baselines} \\
        \hline
        \scriptsize Open CLIP-B/16\ph{+} & \scriptsize 67.5 & \scriptsize 54.5 & \scriptsize 54.2 & \scriptsize 46.8 & \scriptsize 55.8 \\
        \scriptsize OpenAI CLIP-B/16\ph{+} & \scriptsize 67.1 & \scriptsize \textbf{57.6} & \scriptsize 56.5 & \scriptsize 49.3 & \scriptsize 57.6 \\
        \rgray
        \scriptsize \evaTwoclip-B/16\phgray{+} & \scriptsize \textbf{68.6} & \scriptsize 57.4 & \scriptsize \textbf{57.0} & \scriptsize \textbf{50.0} & \scriptsize \textbf{58.3} \\
        \shline
        
        \multicolumn{6}{c}{\scriptsize (b) comparisons with CLIP-\textbf{Large} baselines} \\
        \hline
        \scriptsize Open CLIP-L/14\ph{+} & \scriptsize 73.2 & \scriptsize 58.0 & \scriptsize 58.6 & \scriptsize 50.8 & \scriptsize 60.2 \\
        \scriptsize OpenAI CLIP-L/14\ph{+} & \scriptsize 76.4 & \scriptsize 64.5 & \scriptsize 64.2 & \scriptsize 57.7 & \scriptsize 65.7 \\
        \rgray
        \scriptsize \evaTwoclip-L/14\phgray{+} & \scriptsize \textbf{76.8} & \scriptsize \textbf{65.0} & \scriptsize \textbf{64.9} & \scriptsize \textbf{59.1} & \scriptsize \textbf{66.5} \\
        \shline
        
        \multicolumn{6}{c}{\scriptsize (c) comparisons with \textbf{larger} CLIPs trained with \textbf{more samples}} \\
        \hline
        \scriptsize Open CLIP-H/14\ph{+} & \scriptsize 78.2 & \scriptsize 63.1 & \scriptsize 63.6 & \scriptsize 56.1 & \scriptsize 65.3 \\
        \scriptsize Open CLIP-g/14\ph{+} & \scriptsize 77.8 & \scriptsize 63.9 & \scriptsize 64.1 & \scriptsize 56.9 & \scriptsize 65.7 \\
        \rgray
        \scriptsize \evaOneclip-g/14\phgray{+} & \scriptsize 76.0 & \scriptsize 65.2 & \scriptsize 64.4 & \scriptsize 58.4 & \scriptsize 66.0 \\
        \scriptsize OpenAI CLIP-L/14+ & \scriptsize 78.1 & \scriptsize 64.9 & \scriptsize 65.0 & \scriptsize 58.5 & \scriptsize 66.6 \\
        \rgray
        \scriptsize \evaTwoclip-L/14+ & \scriptsize 78.6 & \scriptsize 65.9 & \scriptsize 66.1 & \scriptsize 60.2 & \scriptsize 67.7 \\
        \scriptsize Open CLIP-G/14\ph{+} & \scriptsize 80.5 & \scriptsize 65.9 & \scriptsize 66.1 & \scriptsize 59.2 & \scriptsize 67.9 \\
        \rgray
        \scriptsize \evaOneclip-g/14+ & \scriptsize 78.9 & \scriptsize 66.7 & \scriptsize 67.0 & \scriptsize 60.9 & \scriptsize 68.4 \\
        \rgray
        \scriptsize \evaTwoclip-E/14\phgray{+} & \scriptsize 82.8 & \scriptsize 68.6 & \scriptsize 68.6 & \scriptsize 62.5 & \scriptsize 70.6 \\
        \rgray
        \scriptsize \evaTwoclip-E/14+ & \scriptsize \textbf{83.1} & \scriptsize \textbf{69.8} & \scriptsize \textbf{69.3} & \scriptsize \textbf{63.4} & \scriptsize \textbf{71.4} \\
    \end{tabular}

\vspace{-1.em}
\caption{\textbf{Summary of \evaclip zero-shot video classification performance.}}
\label{tab: clip zs video cls 4}
\end{table}

\begin{table*}[h]
\vspace{-1.em}
\centering
    \tablestyle{4.2pt}{1.2}
    \begin{tabular}{r|ccc|ccc|ccc|ccc}
        & \multicolumn{6}{c|}{zero-shot \textbf{text} retrieval} & \multicolumn{6}{c}{zero-shot \textbf{image} retrieval} \\
        & \multicolumn{3}{c|}{\scriptsize Flickr30K} & \multicolumn{3}{c|}{\scriptsize COCO} & \multicolumn{3}{c|}{\scriptsize Flickr30K} & \multicolumn{3}{c}{\scriptsize COCO} \\
    
        method{\scriptsize{\ph{+}}} & \scriptsize R@1 & \scriptsize R@5 & \scriptsize R@10 & \scriptsize R@1 & \scriptsize R@5 & \scriptsize R@10 & \scriptsize R@1 & \scriptsize R@5 & \scriptsize R@10 & \scriptsize R@1 & \scriptsize R@5 & \scriptsize R@10 \\
        \shline
        \multicolumn{13}{c}{\scriptsize (a) comparisons with CLIP-\textbf{Base} baselines} \\
        \hline
        \scriptsize OpenAI CLIP-B/16\ph{+} & \scriptsize 81.9 & \scriptsize 96.2 & \scriptsize 98.8 & \scriptsize 52.4 & \scriptsize 76.8 & \scriptsize 84.7 & \scriptsize 62.1 & \scriptsize 85.6 & \scriptsize 91.8 & \scriptsize 33.1 & \scriptsize 58.4 & \scriptsize 69.0 \\
        \scriptsize Open CLIP-B/16\ph{+} & \scriptsize 86.3 & \scriptsize 97.9 & \scriptsize 99.4 & \scriptsize 59.4 & \scriptsize 81.8 & \scriptsize 88.6 & \scriptsize 69.8 & \scriptsize 90.4 & \scriptsize 94.6 & \scriptsize 42.3 & \scriptsize 66.7 & \scriptsize 77.1 \\
        \rgray
        \scriptsize \evaTwoclip-B/16\phgray{+} & \scriptsize \textbf{85.7} & \scriptsize \textbf{96.7} & \scriptsize \textbf{98.9} & \scriptsize \textbf{58.7} & \scriptsize \textbf{80.7} & \scriptsize \textbf{88.2} & \scriptsize \textbf{71.2} & \scriptsize \textbf{91.0} & \scriptsize \textbf{94.7} & \scriptsize \textbf{42.2} & \scriptsize \textbf{66.9} & \scriptsize \textbf{76.3} \\
        \shline
        
        \multicolumn{13}{c}{\scriptsize (b) comparisons with CLIP-\textbf{Large} baselines} \\
        \hline
        \scriptsize OpenAI CLIP-L/14\ph{+} & \scriptsize 85.2 & \scriptsize 97.3 & \scriptsize 99.0 & \scriptsize 56.3 & \scriptsize 79.3 & \scriptsize 86.7 & \scriptsize 65.2 & \scriptsize 87.3 & \scriptsize 92.0 & \scriptsize 36.5 & \scriptsize 61.0 & \scriptsize 71.1 \\
        \scriptsize Open CLIP-L/14\ph{+} & \scriptsize 88.7 & \scriptsize 98.4 & \scriptsize 99.2 & \scriptsize 62.1 & \scriptsize 83.4 & \scriptsize 90.3 & \scriptsize 75.0 & \scriptsize 92.5 & \scriptsize 95.6 & \scriptsize 46.1 & \scriptsize 70.7 & \scriptsize 79.4 \\
        \rgray
        \scriptsize \evaTwoclip-L/14\phgray{+} & \scriptsize {89.7} & \scriptsize {98.6} & \scriptsize {99.2} & \scriptsize {63.7} & \scriptsize {84.3} & \scriptsize {90.4} & \scriptsize {77.3} & \scriptsize {93.6} & \scriptsize {96.8} & \scriptsize {47.5} & \scriptsize {71.2} & \scriptsize {79.7} \\
        \shline
        
        \multicolumn{13}{c}{\scriptsize (c) comparisons with \textbf{larger} CLIPs trained with \textbf{more samples}} \\
        \hline
        
        \scriptsize OpenAI CLIP-L/14+ & \scriptsize 87.4 & \scriptsize 98.3 & \scriptsize 99.3 & \scriptsize 57.9 & \scriptsize 81.2 & \scriptsize 87.9 & \scriptsize 67.3 & \scriptsize 89.0 & \scriptsize 93.3 & \scriptsize 37.1 & \scriptsize 61.6 & \scriptsize 71.5 \\
        \scriptsize Open CLIP-H/14\ph{+} & \scriptsize 90.8 & \scriptsize 99.3 & \scriptsize 99.7 & \scriptsize 66.0 & \scriptsize 86.1 & \scriptsize 91.9 & \scriptsize 77.8 & \scriptsize 94.1 & \scriptsize 96.6 & \scriptsize 49.5 & \scriptsize 73.4 & \scriptsize 81.5 \\
        \scriptsize Open CLIP-g/14\ph{+} & \scriptsize 91.4 & \scriptsize 99.2 & \scriptsize 99.6 & \scriptsize 66.4 & \scriptsize 86.0 & \scriptsize 91.8 & \scriptsize 77.7 & \scriptsize 94.1 & \scriptsize 96.9 & \scriptsize 48.8 & \scriptsize 73.3 & \scriptsize 81.5 \\
        \scriptsize Open CLIP-G/14\ph{+} & \scriptsize 92.9 & \scriptsize 99.3 & \scriptsize 99.8 & \scriptsize 67.3 & \scriptsize 86.9 & \scriptsize 92.6 & \scriptsize \textbf{79.5} & \scriptsize \textbf{95.0} & \scriptsize \textbf{97.1} & \scriptsize \textbf{51.4} & \scriptsize 74.9 & \scriptsize \textbf{83.0} \\
        \rgray
        \scriptsize \evaOneclip-g/14\phgray{+} & \scriptsize 88.3 & \scriptsize 98.3 & \scriptsize 99.3 & \scriptsize 61.8 & \scriptsize 83.3 & \scriptsize 90.0 & \scriptsize 72.6 & \scriptsize 91.6 & \scriptsize 95.1 & \scriptsize 44.1 & \scriptsize 68.5 & \scriptsize 77.3 \\
        \rgray
        \scriptsize \evaOneclip-g/14+ & \scriptsize 91.6 & \scriptsize 99.3 & \scriptsize 99.8 & \scriptsize 68.2 & \scriptsize 87.5 & \scriptsize 92.5 & \scriptsize 78.9 & \scriptsize 94.5 & \scriptsize 96.9 & \scriptsize 50.3 & \scriptsize 74.0 & \scriptsize 82.1 \\
        \rgray
        \scriptsize \evaTwoclip-L/14+ & \scriptsize 89.2 & \scriptsize {98.9} & \scriptsize 99.6 & \scriptsize 64.1 & \scriptsize 85.2 & \scriptsize {90.8} & \scriptsize {77.9} & \scriptsize {94.2} & \scriptsize {96.8} & \scriptsize {47.9} & \scriptsize {71.7} & \scriptsize {80.0} \\
        \rgray
        \scriptsize \evaTwoclip-E/14\phgray{+} & \scriptsize 92.4 & \scriptsize 99.3 & \scriptsize \textbf{99.9} & \scriptsize 68.1 & \scriptsize 87.7 & \scriptsize \textbf{92.8} & \scriptsize 78.8 & \scriptsize 94.6 & \scriptsize 97.0 & \scriptsize 50.8 & \scriptsize 74.7 & \scriptsize 82.5 \\
        \rgray
        \scriptsize \evaTwoclip-E/14+ & \scriptsize \textbf{93.9} & \scriptsize \textbf{99.4} & \scriptsize 99.8 & \scriptsize \textbf{68.8} & \scriptsize \textbf{87.8} & \scriptsize \textbf{92.8} & \scriptsize 78.8 & \scriptsize 94.2 & \scriptsize 96.8 & \scriptsize 51.1 & \scriptsize \textbf{75.0} & \scriptsize 82.7 \\
    \end{tabular}
\vspace{-1.em}
\caption{\textbf{Summary of zero-shot retrival performance on Flickr30K~\cite{flickr30K} and COCO~\cite{lin2014coco}}}
\label{tab: clip zs retrieval}
\end{table*}

\section{Approach}

Training CLIP~\cite{clip} models is very hard and costly. The need for a large batch size and scaling up CLIP models can lead to significant computational resource requirements and even instability training problem. Fortunately, \evaclip offers a highly efficient and effective solution that significantly reduces the computational cost while achieving superior zero-shot performance across a broad range of benchmarks.

\paragraph{Better Initialization}. To improve feature representation and expedite convergence of CLIP models, we adopt pre-trained EVA~\cite{eva, EVA02} models which combines high-level semantics of image-text contrastive learning with geometric and structural capture from masked image modeling. We use pre-trained EVA weights to initialize the image encoder of \evaclip. Our empirical study demonstrates that pre-trained EVA models not only help \evaclip achieve superior performance on various zero-shot benchmarks but also expedites and stabilizes the training process.

\paragraph{Optimizer}. We use the LAMB~\cite{lamb} optimizer for training our \evaclip models. LAMB optimizer is specifically designed for large-batch training, and its adaptive element-wise updating and layer-wise learning rates enhance training efficiency and accelerate convergence rates. Given the exceptionally large batch sizes used for training CLIP models, the original CLIP model uses a batch size of 32,768 and some open-sourced CLIP models even use an extremely large batch size of more than 100k. \evaclip shows that LAMB optimizer is the preferred optimizer for training large-scale CLIP models.

\paragraph{FLIP~\cite{flip}} has shown promising results in large-scale settings. In this work, we leverage FLIP to improve the time efficiency of training CLIP models. Specifically, we randomly mask 50\% image tokens during training esulting in a significant reduction of time complexity by half. This approach also allows for a 2$\times$ increase in batch size, without any additional memory costs.

\section{Experiments}
\paragraph{Settings}. In our experiments, we initialized the vision encoder with pre-trained weights from EVA~\cite{eva, EVA02} and the text encoder with pre-trained weights from either OpenAI CLIP~\cite{clip} or OpenCLIP~\cite{openclip}. 
Specifically, the vision encoder of \evaOneclip is initialized from EVA-01~\cite{eva}, and the vision encoder of \evaTwoclip is initialized from EVA-02~\cite{EVA02}.
We adopt the LAMB optimizer with $\beta_1$ = 0.9, $\beta_2$=0.98, and a weight decay of 0.05. We applied different learning rates and layer decay rates to the vision encoder and text encoder to ensure optimal training. For example, we set the learning rate to 2e-4 for the vision encoder and 2e-5 for the text encoder of \evaOneclip-g during the first 2000 warming-up steps. Afterward, we decayed the learning rate linearly to 0 for the remainder of the training steps. To further improve the training process, we used the $\mathtt{DeepSpeed}$ optimization library~\cite{rasley2020deepspeed} with ZeRO stage-1 optimizer~\cite{rajbhandari2020zero}, gradient checkpointing~\cite{gradcheckpointing} and flash attention~\cite{flashattention} to save memory and accelerate training process. We found that using the $\mathtt{fp16}$ precision with dynamic loss scaling was sufficiently stable throughout the \evaOneclip-g training process, while $\mathtt{bfloat16}$ format was necessary to stabilize the training process of \evaTwoclip-E+.

To construct our training dataset, Merged-2B, we merged 1.6 billion samples from LAION-2B~\cite{laion5b} dataset with 0.4 billion samples from COYO-700M~\cite{kakaobrain2022coyo-700m}.

\paragraph{System-level Comparison}. We present CLIP model configurations and zero-shot accuracies on ImageNet variants and ObjectNet in~\tblref{tab: clip config and results}. \evaTwoclip-E/14+ achieves the highest zero-shot top-1 accuracy of 80.9\% averaged accross all 6 benchmarks, with the smallest performance drop (1.1\% top-1 accuracy gap). Notably, this result surpasses the previous largest and best open-sourced OpenCLIP-G/14~\cite{clipbigg} by 1.9\% on ImageNet and 4.7\% on the average accuracy of 6 benchmarks. Remarkably, with those powerful techniques, the large size \evaTwoclip-L can even reach up to 80.4\% zero-shot top-1 on ImageNet, outperforming OpenCLIP-G/14 with only \app1/6 parameters and \app1/6 image-text training samples.

In ~\tblref{tab: clip zs img cls 27}, we further demonstrate the efficacy and robustness of our approach on all 27 zero-shot image classification benchmarks. Our largest \evaTwoclip-E/14+ achieves averaged 77.5\% on all 27 benchmarks. Notably, our \evaTwoclip-L/14+ model, which only has \app1/2 of the model size and \app1/5 image-text pairs, achieves a 1.2-point averaged improvement over OpenCLIP-H/14.

For video classification, we sample only a single center frame in each video, making it an image classification task. Following the conventional settings, we report the top-1 accuracy for UCF-101~\cite{ucf101} and the mean of top-1 and top-5 accuracy for Kinetics-400~\cite{carreira2017quo}, Kinetics-600~\cite{k600} and Kinetics-700~\cite{k700}. In~\tblref{tab: clip zs video cls 4} we show that \evaclip is also quite effective in zero-shot video recognition benchmarks. 

\tblref{tab: clip zs retrieval} reports the zero-shot image and text retrieval results on Flickr30K~\cite{flickr30K} and COCO~\cite{lin2014coco}. \evaclip outperforms all the competitors at the base and large model size.  While the zero-shot retrieval performance of \evaTwoclip-E/14 is slightly lower than OpenCLIP-G/14, the results are still competitive. We speculate that the main reason is that retrieval tasks depend more on the capacity of the text encoder and the number of training samples, and in comparison, the capacity of the text encoder in \evaTwoclip-E/14 is smaller and the number of training samples is less than that of OpenCLIP-G/14. To this end, we have trained \evaTwoclip-E/14+ with a larger capacity text encoder and more training samples. The results show that this improved model can substantially improve retrieval performance and outperform OpenCLIP-G/14 on zero-shot text retrieval tasks. 

\paragraph{Ablation Study}. We first ablate the \evaclip design in ~\tblref{tab: ablation}. The image encoder is ViT-B/16\cite{vit} model and text encoder is CLIP-B-16. We conducted experiments with a batch size of 32k and evaluated zero-shot accuracy on the ImageNet-1K validation set. It is important to note that we used a shorter training schedule than the final models.

We trained our model on the LAION-400M\cite{laion400m} dataset using the AdamW\cite{Loshchilov2019adamw} optimizer. Compraing to training from scratch, EVA initialization resulted in a 1.8\% increase in zero-shot top-1 accuracy on ImageNet with only \app1/2 seen samples.

Furthermore, we experimented with using the LAMB optimizer instead of AdamW with EVA initialization on the LAION-400M dataset. This resulted in a 0.7\% increase in zero-shot top-1 accuracy on ImageNet with the same seen samples. When 50\% masking was applied, the accuracy decreased by 0.7\% while enjoying a speedup of 2$\times$. These results highlight the significance of LAMB optimizer in training high-performing models and the strategy of masking image tokens in faster training without marginally decreasing accuracy.

 We also conducted experiments with the LAION-2B dataset using EVA initialization, LAMB optimizer, and 50\% masking, which increased the accuracy by 0.7\% compared to LAION-400M. Only half samples were required to achieve the same top-1 accuracy when using the merged-2B dataset. It demonstrates the importance of dataset sizes and the significant convergence speed through merging the two datasets.

%#################################################
% ablation study
%#################################################
\begin{table}[!t]
% \vspace{-1em}
    \centering
    \tablestyle{2pt}{1.2}
    \scriptsize
    \begin{tabular}{l|ccccc|c}
        &   & \evaone &  & mask & samples & {\scriptsize{IN-1K}} \\ 
        method & dataset & init & optimizer &  50\%  & seen & zs top-1 \\
        \shline

         & {\scriptsize{LAION-400M}} & \xmark & AdamW & \xmark & 13B & 67.1 \\
         & \scriptsize{LAION-400M} & \cmark & AdamW & \xmark & 5B & 68.9 \\
        \evaclip-B & \scriptsize{LAION-400M} & \cmark & LAMB & \xmark & 5B & 69.6 \\
         & \scriptsize{LAION-400M} & \cmark & LAMB & \cmark & 5B & 68.9 \\
         & \scriptsize{LAION-2B} & \cmark & LAMB & \cmark & 5B & 69.6 \\
         & \scriptsize{merged-2B} & \cmark & LAMB & \cmark & 2.5B & 69.7 \\

    \end{tabular}
    \vspace{-.5em}
    \caption{\textbf{Ablation Studies}. Our LAION-2B dataset comprises only 1.6B samples.}
    % \vspace{-10pt}
    \label{tab: ablation}
\end{table}

%#################################################
% cost
%#################################################

\begin{table}[!t]
% \vspace{-1em}
    \centering
    \tablestyle{2pt}{1.2}
    \scriptsize
    \begin{tabular}{l|c|c|c|c}
        & mask & flash  \\ 
        method & 50\% & attention & time / 40m samples &  memory / GPU \\
        \shline
        & \xmark & \xmark & 132min & 33GB \\
        & \xmark & \cmark & 76min & 26GB \\
        \evaclip-B & \cmark & \xmark & 64min & 18GB \\
        & \cmark & \cmark & 55min & 16GB \\
    \end{tabular}
    \vspace{-.5em}
    \caption{\textbf{Training time and GPU memory}. Training on 16 NVIDIA 40G-A100 GPUs with the $\mathtt{DeepSpeed}$ ~\cite{rasley2020deepspeed} ZeRO stage-1 optimizer~\cite{rajbhandari2020zero} and gradient checkpointing~\cite{gradcheckpointing}. The batch size is 32k.}
    % \vspace{-10pt}
    \label{tab: computation cost}
\end{table}

\paragraph{Computation Costs}. In \tblref{tab: computation cost}, we present the memory and time cost of our implementation. As shown, masking 50\% of image tokens can accelerate training time by 2$\times$ and using flash atttention can reduce additional 15\% training time.

Using all these techniques, we can train \evaclip with a lower budget than other counterpart CLIP models. For instance, \evaclip-B/16 can be trained on a batch size of 32k and converges within 300 hours using  on 16 NVIDIA 40GB-A100 GPUs. Similarly, the billion-scale EVA CLIP-g/14 can be trained on a batch size of 65k and requires less than 25 days to train 12B samples using 64 NVIDIA 40G-A100 GPUs. These results demonstrate the scalability and effectiveness of our method in achieving state-of-the-art results while maintaining an optimal balance between training time and GPU memory utilization.

% \clearpage

%#################################################
% config
%#################################################

\begin{table}[t!]
    \tablestyle{0.8pt}{1.2}
    \scriptsize
    \begin{tabular}{l|ccc|ccc|ccc}
        & \multicolumn{3}{c|}{image encoder} & \multicolumn{3}{c|}{text encoder} & \multicolumn{3}{c}{\# \scriptsize{params}} \\
        method & layers & width & heads & layers & width & heads & image & text & total \\ 
        \shline
        \evaTwoclip-B/16\ph{+} & 12 &  768 & 12 & 12 &  512 &  8 &  86M &  63M & 149M \\
        \evaTwoclip-L/14\ph{+} & 24 & 1024 & 16 & 12 &  768 & 12 & 304M & 124M & 428M \\
        \evaOneclip-g/14\ph{+} & 40 & 1408 & 16 & 12 & 768 & 12 & 1.0B & 124M & 1.1B \\
        \evaOneclip-g/14+ & 40 & 1408 & 16 & 24 & 1024 & 16 & 1.0B & 354M & 1.3B \\
        \evaTwoclip-E/14\ph{+} & 64 & 1792 & 16 & 24 & 1024 & 16 & 4.4B & 354M & 4.7B \\
        \evaTwoclip-E/14+ & 64 & 1792 & 16 & 32 & 1280 & 20 & 4.4B & 695M & 5.0B \\
    \end{tabular}
    \vspace{-.5em}
    \caption{\textbf{Architecture configurations.}
    }
    \label{evacliparch}
\end{table}

\begin{table}[h]
% \vspace{-.5em}
\centering
\tablestyle{4pt}{1.2}
\scriptsize
\begin{tabular}{l|c}
config & \evaTwoclip-B / -L / -L+ \\
\shline

image enc. weight init. & \eva-B / -L / \evaTwoclip-L \\
text enc. weight init. & OpenAI CLIP-B / -L / \evaTwoclip-L \\

image-text data & Merged-2B \\

image enc. peak learning rate &  2e-4 / 4e-4 / 4e-4 \\
image enc. layer-wise lr decay~\cite{clark2020electra, bao2021beit} & 0.75 / 0.85 / 0.75 \\
text enc. peak learning rate &  2e-5 / 4e-5 / 4e-5 \\
text enc. layer-wise lr decay~\cite{clark2020electra, bao2021beit} & 0.75 / 0.75 / 0.65 \\

learning rate schedule & cosine decay \\

optimizer & LAMB~\cite{lamb} \\
optimizer hyper-parameters & $\beta_1$, $\beta_2$, $\epsilon$ = 0.9, 0.98, 1e-6 \\
weight decay & 0.05 \\

input resolution & 224\suptext{2} / 224\suptext{2} / 336\suptext{2} \\
patch size & 16\suptext{2} / 14\suptext{2} / 14\suptext{2} \\

batch size & 131k / 131k / 61k \\
samples seen & 8B / 4B / 2B \\

drop path~\cite{huang2016deep} & 0.0 \\
random resized crop & (0.9, 1) \\

numerical precision & $\mathtt{DeepSpeed}$ $\mathtt{fp16}$~\cite{rasley2020deepspeed} \\
ZeRO optimizer~\cite{ramesh2021zero} & stage 1 \\

\end{tabular}
\vspace{-.5em}
\caption{\textbf{\evaclip-B and \evaclip-L training setting.}}
% \vspace{-.5em}
\label{tab: clip cfg}
\end{table}

\begin{table}[h]
% \vspace{-.5em}
\centering
\tablestyle{6pt}{1.2}
\scriptsize
\begin{tabular}{l|c}
config & \evaOneclip-g / \evaTwoclip-g+\\
\shline

image enc. weight init. & \evaOne-g \\
text enc. weight init. & Openai CLIP-L / Open CLIP-H \\

image-text data & LAION-400M~\cite{laion400m} / Merged-2B \\

image enc. peak learning rate &  4e-4 \\
image enc. layer-wise lr decay~\cite{clark2020electra, bao2021beit} & 0.85 \\
text enc. peak learning rate &  4e-5 \\
text enc. layer-wise lr decay~\cite{clark2020electra, bao2021beit} & 0.75 \\

learning rate schedule & cosine decay \\

optimizer & AadamW~\cite{adam,Loshchilov2019adamw} / LAMB~\cite{lamb} \\
optimizer hyper-parameters & $\beta_1$, $\beta_2$, $\epsilon$ = 0.9, 0.98, 1e-6 \\
weight decay & 0.05 \\

input resolution & 224\suptext{2} \\
patch size & 14\suptext{2} \\

batch size & 41k / 114k \\
samples seen & 11B / 11B \\

drop path~\cite{huang2016deep} & 0.0 \\
random resized crop & (0.9, 1) \\

numerical precision & $\mathtt{DeepSpeed}$ $\mathtt{fp16}$~\cite{rasley2020deepspeed} \\
ZeRO optimizer~\cite{ramesh2021zero} & stage 1 \\

\end{tabular}
\vspace{-.5em}
\caption{\textbf{\evaclip-g and \evaclip-g+ training setting.}}
% \vspace{-.5em}
\label{tab: clip cfg}
\end{table}

\begin{table}[h]
% \vspace{-.5em}
\centering
\tablestyle{6pt}{1.2}
\scriptsize
\begin{tabular}{l|c}
config & \evaOneclip-E / \evaTwoclip-E+ \\
\shline

image enc. weight init. & \eva-E\\
text enc. weight init. & Open CLIP-H / Open CLIP-G \\

image-text data & LAION-2B~\cite{laion5b} \\

image enc. peak learning rate &  4e-4 \\
image enc. layer-wise lr decay~\cite{clark2020electra, bao2021beit} & 0.9 \\
text enc. peak learning rate &  4e-5 \\
text enc. layer-wise lr decay~\cite{clark2020electra, bao2021beit} & 0.75 \\

learning rate schedule & cosine decay \\

optimizer & LAMB~\cite{lamb} \\
optimizer hyper-parameters & $\beta_1$, $\beta_2$, $\epsilon$ = 0.9, 0.98, 1e-6 \\
weight decay & 0.05 \\

input resolution & 224\suptext{2} \\
patch size & 14\suptext{2} \\

batch size & 115k / 144k \\
samples seen & 4B / 9B \\

drop path~\cite{huang2016deep} & 0.0 \\
random resized crop & (0.9, 1) \\

numerical precision & $\mathtt{DeepSpeed}$ $\mathtt{fp16}$ / $\mathtt{DeepSpeed}$ $\mathtt{bf16}$~\cite{rasley2020deepspeed} \\
ZeRO optimizer~\cite{ramesh2021zero} & stage 1 \\

\end{tabular}
\vspace{-.5em}
\caption{\textbf{\evaclip-E and \evaclip-E+ training setting.}}
% \vspace{-.5em}
\label{tab: clip cfg}
\end{table}

%%%%%%%% REFERENCES
{
\fontsize{8.2pt}{9.84pt}\selectfont
\bibliographystyle{ieee_fullname}
\bibliography{evaclip}
}

\end{document}